%% file: BaSSL.tex
\documentclass{article} 
\usepackage{iclr2022_conference,times}

\input{math_commands.tex}

\usepackage{hyperref}
\usepackage{url}

\usepackage{graphicx}
\usepackage{amsmath,amssymb} 
\usepackage{makecell}
\usepackage{color}
\usepackage{xcolor,colortbl}
\usepackage{enumitem}
\usepackage{booktabs}
\usepackage{float}
\usepackage{tabulary,multirow,overpic,xcolor}
\usepackage{verbatim}
\usepackage{enumitem}
\usepackage[caption=false]{subfig}
\usepackage{array}
\usepackage{arydshln}
\usepackage{adjustbox}
\usepackage{siunitx}
\usepackage{algorithm}
\usepackage{algpseudocode}
\usepackage{hhline}
\usepackage{tabu}
\usepackage{wrapfig}
\usepackage{lipsum}
\usepackage{listings}

\newcommand{\ie}{\textit{i}.\textit{e}.}
\newcommand{\eg}{\textit{e}.\textit{g}.}

\newcommand{\ma}[1]{\mathbf{#1}} 

\newcolumntype{x}[1]{>{\centering\arraybackslash}p{#1pt}}

\title{Boundary-aware Self-Supervised Learning for Video Scene Segmentation}

\author{%
Jonghwan Mun$^{*}$, Minchul Shin\thanks{Equal contribution ~~$\dagger$ Corresponding authors}~~, Gunsoo Han \\
Kakao Brain \\
\texttt{\{jason.mun,craig.starr,coco.han\}@kakaobrain.com} \\
\AND
Sangho Lee, Seongsu Ha, Joonseok Lee$^{\dagger}$ \\
Seoul National University \\
\texttt{\{sangho.lee,sha17,joonseok\}@snu.ac.kr} \\
\And
Eun-Sol Kim$^{\dagger}$\\
Hanyang University \\
\texttt{eunsolkim@hanyang.ac.kr}
}

\iclrfinalcopy 
\begin{document}
\maketitle

\begin{abstract}
Self-supervised learning has drawn attention through its effectiveness in learning in-domain representations with no ground-truth annotations; in particular, it is shown that properly designed pretext tasks (e.g., contrastive prediction task) bring significant performance gains for downstream tasks (e.g., classification task). Inspired from this, we tackle video scene segmentation, which is a task of temporally localizing scene boundaries in a video, with a self-supervised learning framework where we mainly focus on designing effective pretext tasks. In our framework, we discover a pseudo-boundary from a sequence of shots by splitting it into two continuous, non-overlapping sub-sequences and leverage the pseudo-boundary to facilitate the pre-training. Based on this, we introduce three novel boundary-aware pretext tasks: 1) Shot-Scene Matching (SSM), 2) Contextual Group Matching (CGM) and 3) Pseudo-boundary Prediction (PP); SSM and CGM guide the model to maximize intra-scene similarity and inter-scene discrimination while PP encourages the model to identify transitional moments. Through comprehensive analysis, we empirically show that pre-training and transferring contextual representation are both critical to improving the video scene segmentation performance. Lastly, we achieve the new state-of-the-art on the MovieNet-SSeg benchmark. The code is available at \url{https://github.com/kakaobrain/bassl}.
\end{abstract}

\section{Introduction}

Understanding long videos such as movies, for an AI system, has been viewed as an extremely challenging task.
In contrast, as cognitive science~\citep{tversky2013event} tells us, for humans, it is naturally achieved by breaking down a video into meaningful units (\eg, event) and reasoning about these units and their relation~\citep{GEBD}.
From this point of view, dividing a long video into a series of shorter temporal segments can be considered as an essential step towards the high-level video understanding.
Motivated by this, in this paper, we tackle the video scene segmentation task, temporally localizing scene boundaries from a long video;
the term scene is widely used in filmmaking, where the scene is considered as a basic unit for understanding the story of movies and is composed of a series of semantically cohesive shots.

\begin{figure}[!t]
	\centering
	\scalebox{1}{
		\includegraphics[width=\linewidth]{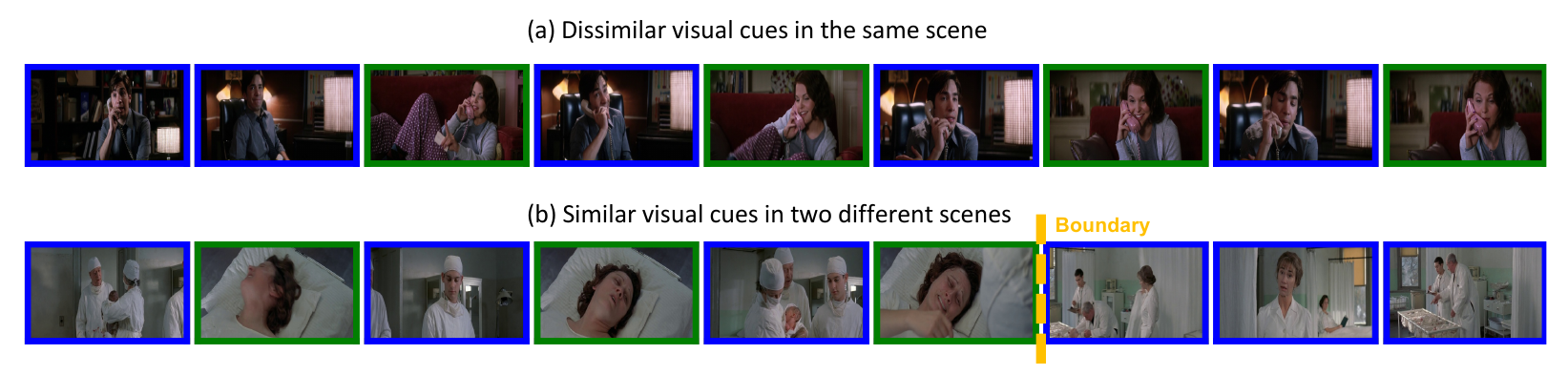}
	}
	\vspace{-0.8cm}
	\caption{
		Examples of the video scene segmentation. In each row, we visualize the shots including similar visual cues (\eg, characters, places, etc.) with the same colored border.
	}
	\vspace{-0.2cm}
	\label{fig:scene_example}
\end{figure}

One of the biggest challenges with video scene segmentation is that it is not achieved simply by detecting changes in visual cues.
As shown in Figure~\ref{fig:scene_example}(a), we present an example of nine shots, all of which belong to the same scene, where two characters are talking on the phone; the overall visual cues within the scene do not stay the same but rather change repeatedly when each character appears.
On the other hand, Figure~\ref{fig:scene_example}(b) shows two different scenes which contain visually similar shots (highlighted in blue) where the same character appears in the same place. 
Thus, it is expected that two adjacent scenes which share shots with similar visual cues need to be contextually discriminated.
From this observation, it is important for the video scene segmentation task to model contextual relation between shots by maximizing 1) \textit{intra-scene similarity} (\ie, the shots in the same scene should be close to each other), and 2) \textit{inter-scene discrimination} across two adjacent scenes (\ie, two neighbor shots across the scene boundary should be distinguishable).

Supervised learning approaches (\eg, \cite{lgss}) are clearly limited due to the lack of large-scale datasets with reliable ground-truth annotations.
Recently, self-supervision~\citep{chen2020simple,caron2020unsupervised,he2020momentum,SCRL} is spotlighted through its effectiveness in learning in-domain representation without relying on costly ground-truth annotations.
The self-supervised learning methods \citep{shotcol,feichtenhofer2021large,TCLR,CVRL} in the video domain are often designed to learn spatio-temporal patterns in short clips (\eg, shots in movies).
This kind of learned representation is generic and can be applied to many video understanding tasks (\eg, action classification).
However, shot-level representation is insufficient for video scene segmentation because this task requires not only a good representation for individual shots but also contextual representation considering neighboring shots at a higher level as observed in Figure~\ref{fig:scene_example}.
Motivated by this, we set our main goal to design self-supervised objectives (\ie, pretext tasks) that maximize intra-scene similarity as well as discriminate shots from different scenes by taking contextual information into account.
This raises a penetrating question: \textit{how can we design boundary-relevant pretext tasks without access to ground truth boundary annotations?}

We introduce a novel \textbf{B}oundary-\textbf{a}ware 
\textbf{S}elf-\textbf{S}upervised \textbf{L}earning (BaSSL) framework, which leverages pseudo-boundaries to learn contextual representation effective in capturing semantic transition during pre-training stage, thus leading to precise scene boundary detection.
The pseudo-boundary is obtained by dividing the input sequence of shots into two semantically disjoint sub-sequences, and we use it to define boundary-relevant pretext tasks that are beneficial to the video scene segmentation task.
On top of the discovered two sub-sequences and a pseudo-boundary, three boundary-aware pretext tasks are proposed: 1) Shot-Scene Matching (SSM); 2) Contextual Group Matching (CGM); and 3) Pseudo-boundary Prediction (PP).
Note that SSM and CGM encourage the model to maximize intra-scene similarity and inter-scene discrimination, while PP enables the model to learn the capability of identifying transitional moments.
In addition, we perform Masked Shot Modeling (MSM) task inspired by \cite{cbt} to further learn temporal relationship between shots.
The comprehensive analysis demonstrates the effectiveness of the proposed framework (\ie, pre-training of contextual relationship between shots) as well as the contribution of the proposed individual components (\ie, the algorithm for pseudo-boundary discovery and boundary-aware pretext tasks).

Our main contributions are summarized as follows:
($i$) we introduce a novel boundary-aware pre-training framework which leverages pseudo-boundaries to learn contextual relationship between shots during the pre-training;
($ii$) we propose three boundary-aware pretext tasks, which are carefully designed to learn essential capabilities required for the video scene segmentation task;
($iii$) we perform extensive ablations to demonstrate the effectiveness of the proposed framework, including the observation that our framework is complementary to the existing framework;
($iv$) we achieve the new state-of-the-art on the MovieNet-SSeg benchmark while outperforming existing self-supervised learning-based methods by large margins.

\section{Related Work}

\paragraph{Video scene segmentation}
approaches formulate the task as a problem of temporal grouping of shots.
In this formulation, the optimal grouping can be achieved by clustering-based~\citep{rui1998exploring,rasheed2003scene,GraphCut,SCSA}, dynamic programming-based~\citep{DP,storygraph,grouping} or multi-modal input-based~\citep{liang2009novel,sidiropoulos2011temporal} methods.
However, the aforementioned methods have been trained and evaluated on small-scale datasets such as OVSD~\citep{OVSD} and BBC~\citep{siamese} which can produce a poorly generalized model.
Recently, \cite{movienet} introduce a large-scale video scene segmentation dataset (\ie, MovieNet-SSeg) that contains hundreds of movies.
Training with large-scale data, \cite{lgss} proposes a strong supervised baseline model that performs a shot-level binary classification followed by grouping using the prediction scores. In addition, \cite{shotcol} proposes a shot contrastive pre-training method that learns shot-level representation.
We found ShotCoL~\citep{shotcol} to be the most similar work to our method. However, our method is different from ShotCoL in that we specifically focus on learning contextual representations by considering the relationship between shots. We refer interested readers to the supplementary material for a more detailed comparison with ShotCoL.

\paragraph{Action segmentation in videos}
is one of the related works for video scene segmentation, which identifies action labels of individual frames, thus can divide a video into a series of action segments.
Supervised methods~\citep{lea2016segmental,farha2019ms} proposed CNN-based architectures to effectively capture temporal relationship between frames in order to address an over-segmentation issue.
As frame-level annotations are prohibitively costly to acquire, weakly supervised methods~\citep{chang2019d3tw,li2019weakly,li2020set,souri2021fast,shen2021learning,zhukov2019cross,fried2020learning} have been suggested to use an ordered list of actions occurring in a video as supervision.
Most of the methods are trained to find (temporal) semantic alignment between frames and a given action list using an HMM-based architecture~\citep{kuehne2018hybrid}, a DP-based assignment algorithm~\citep{fried2020learning} or a DTW-based temporal alignment method~\citep{chang2019d3tw}.
Recently, unsupervised methods~\citep{kumar2021unsupervised,wang2021unsupervised,kukleva2019unsupervised,li2021action,vidalmata2021joint} have been further proposed;
in a nutshell, clustering-based prototypes are discovered from unlabeled videos, then the methods segment the videos by assigning prototypes (corresponding to one of the actions) into frames.
In contrast to the action segmentation task that is limited to localizing segments each of which represents a single action within an activity, video scene segmentation requires localizing more complex segments each of which may be composed of more than two actions (or activities).

\paragraph{Self-supervised learning in videos}
has been actively studied for the recent years with approaches proposing various pretext tasks such as future frame prediction~\citep{srivastava2015unsupervised,vondrick2016generating,ahsan2018discrimnet}, temporal ordering of frames~\citep{misra2016shuffle,lee2017unsupervised,xu2019self}, geometric transformations prediction~\citep{jing2018self}, colorization of videos~\citep{vondrick2018tracking} and contrastive prediction~\citep{feichtenhofer2021large,CVRL,TCLR}.
In addition, CBT~\citep{cbt,videobert} proposes a pretext task of masked frame modeling to learn temporal dependency between frames (or clips).
Note that since most of those methods are proposed for the classification task, they would be sub-optimal to the video scene segmentation task.
On the other hand, BSP~\citep{bsp} proposes boundary-sensitive pre-text tasks based on synthesized pseudo-boundaries that are obtained by concatenating two clips sampled from different videos.
However, strictly speaking, BSP is not a self-supervised learning algorithm since it requires video-level class labels to synthesize pseudo-boundaries;
the proposed pretext tasks are not applicable to videos such as movies that are hard to define semantic labels.
Also, note that we empirically show that pseudo-boundaries identified by our method are more effective for pre-training than synthesized pseudo-boundaries.

\section{Boundary-aware Self-supervised Learning (BaSSL)}

In this section, we introduce our proposed approach, Boundary-aware Self-Supervised Learning (BaSSL). We start with the problem formulation followed by the overview. Then, we describe our novel boundary-aware pretext tasks for pre-training.


\subsection{Problem Formulation}

\paragraph{Terminologies}
A video (\eg, documentaries, TV episodes and movies) is assumed to have a hierarchical structure at three-level semantics: scene, shot and frame.
In detail, a video is a sequence of scenes, defined as a semantic unit for making a story. A scene is a series of shots, which is a set of frames physically captured by the same camera during an uninterrupted period of time.

\paragraph{Video Scene Semantic Segmentation Task}
Given a video, which contains a series of $N$ shots $\{\mathbf{s}_1,...,\mathbf{s}_N\}$ with class labels $\{y_1,...,y_N\}$ where $y_i \in \{0,1\}$ indicating if it is at the scene boundary (more precisely, if it is the last shot of a scene), the video scene segmentation task is formulated as a simple binary classification problem at individual shot level.
By definition, a scene boundary is where the semantic of a shot is considerably different from its (one-way) neighbors.
Thus, it is in nature important to capture and leverage contextual transition across the scenes.
Consequently, it is a common practice that the information of the neighbor shots are leveraged together when determining scene boundaries.
With this formulation, existing approaches~\citep{lgss, shotcol} adopt a sliding window scheme with a window $\ma{S}_n=\{ \mathbf{s}_{n-K}, ..., \mathbf{s}_n, ..., \mathbf{s}_{n+K} \}$ containing a sequence of $2K+1$ shots centered at the $n^{\text{th}}$ shot, $\mathbf{s}_n$, where $K$ is the number of neighbor shots before and after $\mathbf{s}_n$.
Then, the supervised learning methods typically train a parameterized ($\theta$) model by maximizing the expected log-likelihood:
\begin{equation*}
    \theta^{*} = \argmax_{\theta} \mathbb{E} \left[ \log p_{\theta} (y_n|\ma{S}_n) \right].
    \label{eq:sbd}
\end{equation*}
 Note that each shot $\mathbf{s}$ is given by a set of $N_k$ key-frames, resulting in a tensor with size of $(N_k, C, H, W)$ where $C$, $H$ and $W$ are the RGB channels, the height and the width, respectively.

\begin{figure}[!t]
	\centering
	\scalebox{1}{
		\includegraphics[width=\linewidth]{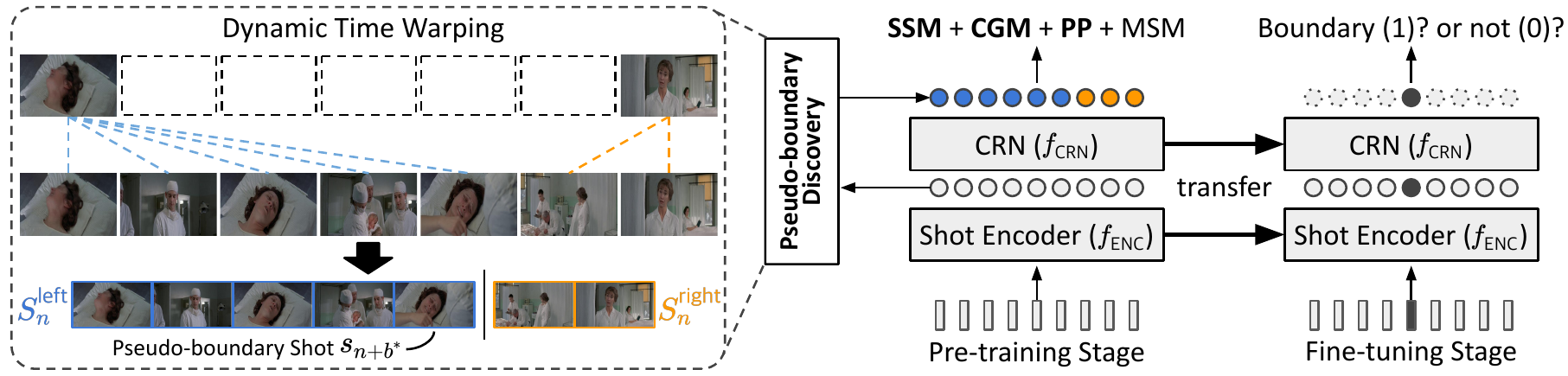}
	}
	\vspace{-0.7cm}
	\caption{
		Overall pipeline of our proposed framework, BaSSL.
	}
	\label{fig:framework}
\end{figure}

\begin{figure}[!t]
	\centering
	\scalebox{1}{
		\includegraphics[width=\linewidth]{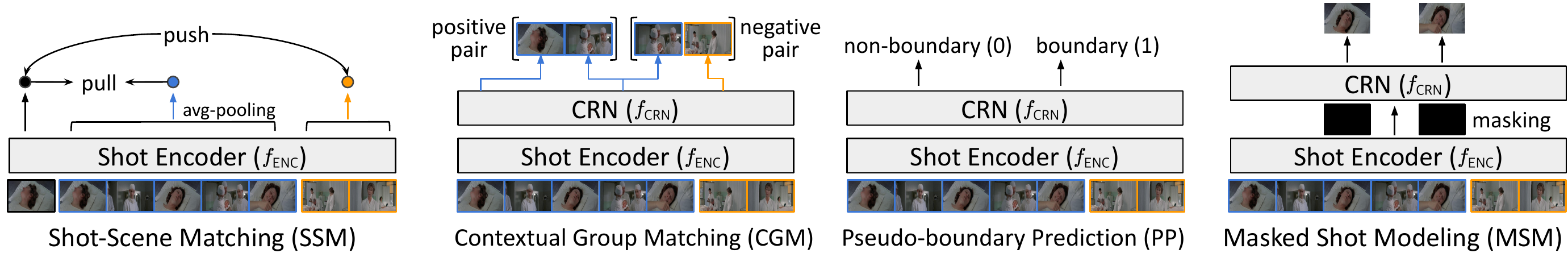}
	}
	\vspace{-0.7cm}
	\caption{
		Illustration of four pre-training pretext tasks.
	}
	\label{fig:pretext_tasks}
\end{figure}

\subsection{Model Overview}
\label{sec:method:overview}

Our method is based on two-stage training following common practice~\citep{shotcol}: pre-training on large-scale unlabeled data with self-supervision and fine-tuning on relatively small labeled data via transfer learning. Our main focus is in the pre-training stage, aiming at designing effective pretext tasks for video scene segmentation.

As illustrated in Figure~\ref{fig:framework}, the model ($\theta$) consists of two main components: 1) \emph{shot encoder} embedding a shot by capturing its spatio-temporal patterns, and 2) \emph{contextual relation network} (CRN) capturing relationship between shots.
Given a sequence $\ma{S}_n=\{\mathbf{s}_{n-K}, ..., \mathbf{s}_n, ..., \mathbf{s}_{n+K}\}$ of shots centered at $\mathbf{s}_n$, two-level representations are extracted as follows:
\begin{equation*}
    \mathbf{e}_n = f_{\text{ENC}}(\mathbf{s}_n) 
    ~~\text{and}~~
    \ma{C}_n = f_{\text{CRN}}(\ma{E}_n),
\end{equation*}
where 
$f_{\text{ENC}}\colon\mathbb{R}^{N_k\times C\times H\times W}\hspace{-0.15cm}\rightarrow\hspace{-0.1cm} \mathbb{R}^{D_e}$ and 
$f_{\text{CRN}}\colon\mathbb{R}^{(2K+1)\times D_e}\hspace{-0.15cm}\rightarrow\hspace{-0.1cm} \mathbb{R}^{(2K+1)\times D_c}$ represent the shot encoder and the contextual relation network while $D_e$ and $D_c$ mean dimensions of encoded and contextualized features, respectively.
$\mathbf{e}_n$ is an encoding of shot $\mathbf{s}_n$ by $f_{\text{ENC}}$ while $\ma{E}_n = \{ \mathbf{e}_{n-K}, ..., \mathbf{e}_{n}, ..., \mathbf{e}_{n+K}\}$ and $\ma{C}_n = \{ \mathbf{c}_{n-K}, ..., \mathbf{c}_{n}, ..., \mathbf{c}_{n+K}\}$ correspond to the input and output feature sequence for $f_{\text{CRN}}$, respectively.

On top of encoded shot representations, BaSSL extracts a pseudo-boundary (left box of Figure~\ref{fig:framework}) to self-supervise the model instead of relying on ground-truth annotations. 
To be specific, we leverage the dynamic time warping technique to divide the input sequence of shots into two semantically disjoint sub-sequences and output a pseudo-boundary. (See Section~\ref{sec:method:dtw} for more details.)

Then, as presented in Figure~\ref{fig:pretext_tasks}, using the discovered pseudo-boundary, we devise three novel boundary-aware pretext tasks in Section~\ref{sec:method:losses}: 1) \emph{Shot-Scene Matching} to match shots with their associated scenes, 2) \emph{Contextual Group Matching} to align shots whether they belong to the same scene or not and 3) \emph{Pseudo-boundary Prediction} to capture semantic changes.
In addition, we adopt the masked shot modeling in CBT~\citep{cbt} to further learn temporal relationship between shots.
After pre-trained with the four pretext tasks, the model is fine-tuned with labeled video scene segmentation data. (See Section~\ref{sec:finetuning} for more details.) 

\begin{figure}[!t]
	\centering
	\scalebox{1}{
		\includegraphics[width=\linewidth]{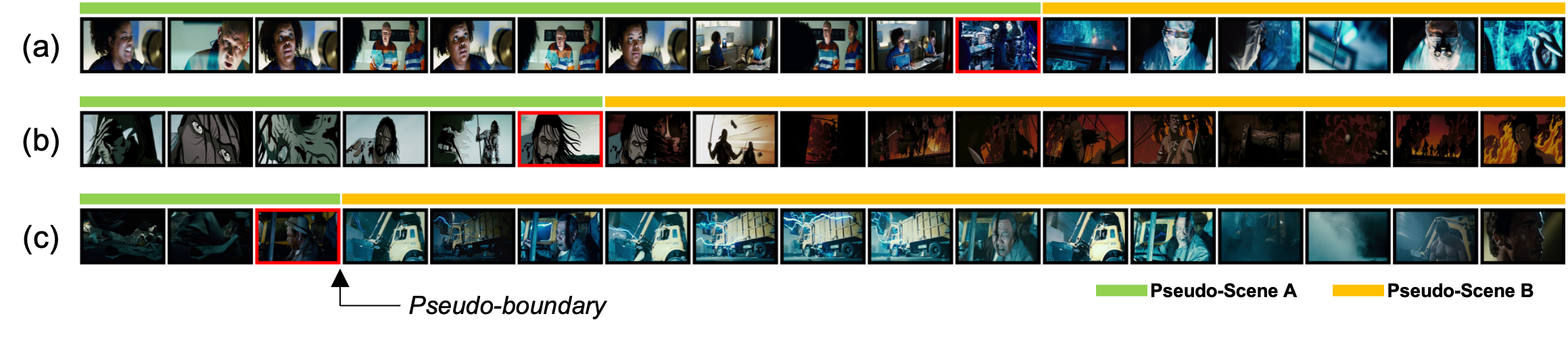}
	}
	\vspace{-0.8cm}
	\caption{
		An example in each row shows a sequence of shots sampled from the same scene where there exists no ground-truth scene-level boundary.
		Our method finds a pseudo-boundary shot (highlighted in red) that divides a sequence into two pseudo-scenes (represented by green and orange bars, respectively) so that semantics (\eg, places, characters) maximally changes.
	}
	\vspace{-0.3cm}
	\label{fig:pseudo_boundary}
\end{figure}

\subsection{Pseudo-boundary Discovery}
\label{sec:method:dtw}

The goal of our pre-training is to make a model effective in capturing the semantic transition, thus leading to higher performance in downstream task (\ie, video scene segmentation).
For this purpose, we leverage a pseudo-boundary---a probable moment where the actual semantic transition occurs---as a clue for self-supervision.
Given an input sequence, we simply find a single moment (or boundary) with the maximum semantic transition.
Note that even a sequence of shots without strong scene-level semantic transition, there always exists a shot within the sequence across which the semantic transition is maximum, and we use this shot as a pseudo-boundary.
Figure~\ref{fig:pseudo_boundary} shows examples where we intentionally infer a pseudo-boundary based on a sequence of shots sampled from the same scene (that is, no scene boundary exists according to the ground truth);
we observe that the resulting two sub-sequences are cognitively distinguishable.

The process, dividing the input sequence $\ma{S}_n$ into two continuous, non-overlapping sub-sequences $\ma{S}^\text{left}_n$ and $\ma{S}^\text{right}_n$ with maximum semantic transition, can be seen as a temporal alignment problem between $\ma{S}_n$ and $\ma{S}_n^\text{slow}$;
specifically, observing the first shot should belong to $\ma{S}^\text{left}_n$ and the last one to $\ma{S}^\text{right}_n$, we define $\ma{S}^\text{slow}_n = \{\mathbf{s}_{n-K}, \mathbf{s}_{n+K}\}$, which can be seen as a same video with $\ma{S}_n$ with lower sampling frequency. 
Then, the problem becomes aligning intermediate shots either to the first shot $\mathbf{s}_{n-K}$ or the last shot $\mathbf{s}_{n+K}$ while preserving continuity.

Under the problem setting, we adopt dynamic time warping (DTW) \citep{DTW} to find the optimal alignment between $\ma{S}_n$ and $\ma{S}^\text{slow}_n$. In detail, DTW solves the following optimization problem using dynamic programming to maximize semantic coherence of the resulting two sub-sequences among all possible boundary candidates:
\begin{align*}
    b^* = \argmax_{b=-K+1,...,K-1} \frac{1}{b+K} \sum_{i=-K+1}^b \text{sim}(\mathbf{e}_{n-K}, \mathbf{e}_{n+i}) + \frac{1}{K-b-1} \sum_{j = b+1}^{K-1} \text{sim}(\mathbf{e}_{n+K}, \mathbf{e}_{n+j}),
\end{align*}
where $b$ is the candidate boundary offset, $b^*$ is the optimal boundary offset, and $\text{sim}(\mathbf{x}, \mathbf{y}) = \frac{\mathbf{x}^\top \mathbf{y}}{\lVert \mathbf{x}\rVert \lVert \mathbf{y}\rVert}$ computes cosine similarity between encodings of the given two shots. Two sub-sequences are inferred as $\ma{S}_n^{\text{left}} = \{\mathbf{s}_{n-K}, ..., \mathbf{s}_{n+b^*}\}$ and $\ma{S}_n^{\text{right}} = \{\mathbf{s}_{n+b^*+1}, ..., \mathbf{s}_{n+K}\}$. $\mathbf{s}_{n+b^*}$ is the pseudo-boundary shot, which is the last shot of $\ma{S}_n^{\text{left}}$.
More examples of pseudo-boundaries identified by our algorithm is presented in Figure~\ref{fig:example_gt} of the supplementary material.
The results are used for learning boundary-aware pretext tasks, which will be described below.

\subsection{Pre-training Objectives}
\label{sec:method:losses}

As illustrated in Figure~\ref{fig:pretext_tasks}, we propose three novel boundary-aware pretext tasks---1) shot-scene matching, 2) contextual group matching and 3) pseudo-boundary prediction---and adopt an additional standard pretext task (\ie, masked shot modeling). 

\paragraph{Shot-Scene Matching (SSM)}
The objective of this task is to make the representations of a shot and its associated scene similar to each other, while the representations of the shot and other scenes dissimilar.
In other words, SSM encourages the model to maximize intra-scene similarity, while minimizing inter-scene similarity.
Considering the splitted two sub-sequences ($\ma{S}_n^{\text{left}}$ and $\ma{S}_n^{\text{right}}$) as pseudo-scenes, we train the model using the InfoNCE loss~\citep{infoNCE}:
\begin{align}
    \mathcal{L}_{\text{SSM}} &= \mathcal{L}_\text{NCE} \left( h_{\text{SSM}} (\mathbf{e}_{n-K}), h_{\text{SSM}} (\mathbf{r}(\ma{S}_n^{\text{left}})) \right) 
      + \mathcal{L}_ \text{NCE} \left( h_{\text{SSM}} (\mathbf{e}_{n+K}),
      h_{\text{SSM}} (\mathbf{r}(\ma{S}_n^{\text{right}})) \right), \nonumber  \\
    \mathcal{L}_\text{NCE}(\mathbf{e}, \mathbf{r}) &= 
        -\log \frac{ e^{\text{sim}(\mathbf{e}, \mathbf{r}) / \tau}}{
          e^{ \text{sim}( \mathbf{e}, \mathbf{r}) / \tau} 
            + \sum_{\bar{ \mathbf{e}} \in \mathcal{N}_e} e^{ \text{sim}(\bar{\mathbf{e}}, \mathbf{r}) / \tau}
            + \sum_{\bar{\mathbf{r}} \in \mathcal{N}_r} e^{ \text{sim}(\mathbf{e}, \bar{\mathbf{r}}) / \tau} }, \label{eq:nce_loss}
\end{align}
where $h_{\text{SSM}}$ is a SSM head of a linear layer, $\tau$ is a temperature hyperparameter and $\mathbf{r}(\mathbf{S})$ means a scene-level representation; we use the averaged encoding of shots in the sub-sequence $\mathbf{S}$.
$\mathcal{N}_e$ and $\mathcal{N}_r$ in Eq.~(\ref{eq:nce_loss}) are constructed using other shots and pseudo-scenes in a mini-batch, respectively.

\paragraph{Contextual Group Matching (CGM)}
Since directly matching representations of shots and scenes would not be effective when the scenes are composed of visually dissimilar shots, CGM is introduced to bridge this gap. 
Similar to SSM, CGM is also designed to maximize intra-scene similarity and inter-scene discrimination.
However, CGM measures semantic coherence of the shots rather than comparing visual cues.
With CGM, the model learns to decide if the given two shots belong to the same group (\ie, scene) or not.
In detail, we use the center shot $\mathbf{s}_n$ in the input sequence as the anchor and construct a triplet of ($\mathbf{s}_n$,~$\mathbf{s}_{\text{pos}}$,~$\mathbf{s}_{\text{neg}}$).
We sample each shot from $\ma{S}_n^{\text{left}}$ and $\ma{S}_n^{\text{right}}$; the one sampled within the same sub-sequence with $\mathbf{s}_n$ is used as the positive shot $\mathbf{s}_{\text{pos}}$, while the other as the negative $\mathbf{s}_{\text{neg}}$.
The CGM loss is defined using a binary cross-entropy loss as follows:
\begin{equation*}
    \mathcal{L}_{\text{CGM}} = - \log \left( h_{\text{CGM}} (\mathbf{c}_n, \mathbf{c}_{\text{pos}}) \right) - \log \left( 1 - h_{\text{CGM}}( \mathbf{c}_n, \mathbf{c}_{\text{neg}}) \right),
\end{equation*}
where $h_{\text{CGM}}$ is a CGM head taking two shots as input and predicting a matching score.
$\mathbf{c}_n$, $\mathbf{c}_{\text{pos}}$ and $\mathbf{c}_{\text{neg}}$ are the contextualized features by $f_\text{CRN}$ for the center, positive and negative shots, respectively.

\paragraph{Pseudo-boundary Prediction (PP)}
Through the above two pretext tasks, our model learns the contextual relationship between shots.
In addition to these, we design an extra pretext task, PP, which is more directly related to boundary detection;
PP makes the model have a capability of identifying transitional moments that semantic changes.
Based on the pseudo-boundary shot and one randomly sampled non-boundary shot, the PP loss is defined as a binary cross-entropy loss:
\begin{equation*}
    \mathcal{L}_{\text{PP}} = - \log \left( h_{\text{PP}} (\mathbf{c}_{n+b^{*}}) \right) - \log \left( 1 - h_{\text{PP}} (\mathbf{c}_{\bar{b}}) \right),
\end{equation*}
where $h_{\text{PP}}$ is a PP head that projects the contextualized shot representation to a probability distribution over binary class.
$\mathbf{c}_{n+b^{*}}$ and $\mathbf{c}_{\bar{b}}$ indicate the contextualized representation from $f_\text{CRN}$ for the pseudo-boundary shot $\mathbf{s}_{n+b^*}$ and randomly sampled non-boundary shot $\mathbf{s}_{\bar{b}}$, respectively.

\paragraph{Masked Shot Modeling (MSM)}
Inspired by masked frame modeling \citep{cbt,videobert}, we adopt the MSM task whose goal is to reconstruct the representation of masked shots based on the their surrounding shots.
In this task, given a set of encoded shot representations, we randomly apply masking each of them with a probability of 15\%.
For a set $\mathcal{M}$ of masked shot offsets, we learn to regress the output on each masked shot to its encoded shot representation, which is given by
\begin{align}
    \mathcal{L}_{\text{MSM}} &= \sum_{m \in \mathcal{M}} \lVert \mathbf{e}_{m} - h_{\text{MSM}}(\mathbf{c}_{m}) \rVert_2^2,
\end{align}
where $h_{\text{MSM}}$ is a MSM head to match the dimension of contextualized shot representation with that of encoded one.
$\mathbf{e}_{m}$ and $\mathbf{c}_{m}$ denote the encoded representation by $f_\text{ENC}$ and contextualized representation by $f_\text{CRN}$ for a masked shot $\mathbf{s}_m$, respectively.

\paragraph{Pre-training loss}
The final pre-training loss is defined by
\begin{equation*}
    \mathcal{L}_{\text{pretrain}} = \alpha_1\mathcal{L}_{\text{SSM}} + \alpha_2\mathcal{L}_{\text{CGM}} + \alpha_3\mathcal{L}_{\text{PP}} + \alpha_4\mathcal{L}_{\text{MSM}}, 
\end{equation*}
where $\alpha_1, \alpha_2, \alpha_3$, and $\alpha_4$ are hyperparameters to balance the pretext tasks while all are set to 1.

\subsection{Fine-tuning for Scene Boundary Detection}
\label{sec:finetuning}
Recall that we formulate the video scene segmentation as a binary classification task to identify contextual transition across the scene. 
Different from the pre-training stage, given an input sequence of shots $\ma{S}_n$, we employ a scene boundary detection head $h_{\text{SBD}}$ to infer a prediction from the contextualized representation ($\mathbf{c}_n$) for the center shot $\mathbf{s}_n$.
Following \cite{shotcol}, we freeze the parameters of the shot encoder and then train only the contextual relation network and the scene boundary detection head using a binary cross-entropy loss with the ground truth label $y_n$ as follows:
\begin{equation*}
    \mathcal{L}_{\text{finetune}} = -y_n\log (h_{\text{SBD}}(\mathbf{c}_n) ) + (1-y_n)\log (1-h_{\text{sbd}}(\mathbf{c}_n) ).
\end{equation*}
Note that, with a sidling window scheme, individual shots are decided to be a scene boundary when its prediction score is higher than a pre-defined threshold (set to 0.5).

\section{Experiment}

\begin{table*}[t]
    \caption{
        Comparison with other algorithms.
        $\dagger$ and $\ddagger$ denote that the numbers are copied from \citep{lgss} and \citep{movienet}, respectively.
        $^\star$ indicates the methods exploiting additional modalities or semantics (\eg, audio, place, cast, transcript).
        The best numbers are in bold.
    }
    \vspace{0.1cm}
    \centering
    \scalebox{0.9}{
        \begin{tabular}{l|cccc}
            \toprule
            Method & AP ($\uparrow$) & mIoU ($\uparrow$) & AUC-ROC ($\uparrow$) & F1 ($\uparrow$) \\
            \hline \hline
            
            \rowcolor[gray]{0.85}\multicolumn{5}{l}{\textit{\textbf{Supervised Learning}}} \\ 
            \hline
            Siamese~\citep{siamese}$\ddagger$ & 35.80 & 39.60 & - & - \\
            MS-LSTM~\citep{movienet}$\ddagger^\star$ & 46.50 & 46.20 & - & - \\
            LGSS~\citep{lgss}$\dagger^\star$ & 47.10 & 48.80 & - & - \\
            
            \hline\hline
            \rowcolor[gray]{0.85}\multicolumn{5}{l}{\textit{\textbf{Unsupervised Learning}}} \\ 
            \hline
            GraphCut~\citep{GraphCut}$\dagger$ & 14.10 & 29.70 & & - \\
            SCSA~\citep{SCSA}$\dagger$ & 14.70 & 30.50 & - & - \\
            DP~\citep{DP}$\dagger$ & 15.50 & 32.00 & - & - \\
            Story Graph~\citep{storygraph}$\dagger$ & 25.10 & 35.70 & - & - \\
            Grouping~\citep{grouping}$\ddagger$$^\star$ & 33.60 & 37.20 & & - \\
            BaSSL w/o fine-tuning (10 epochs) & 31.55 & 39.36 & 71.67 & 32.55 \\ 
            
            \hline\hline
            \rowcolor[gray]{0.85}\multicolumn{5}{l}{\textit{\textbf{Self-supervised Learning}}} \\
            \hline
            ShotCoL~\citep{shotcol} & 53.40 & - & - & - \\
            
            BaSSL (10 epochs)  & 56.26 $\pm$0.04 & 49.50 $\pm$0.11 & 90.27 $\pm$0.02 & 45.70 $\pm$0.24 \\
            BaSSL (40 epochs) & \textbf{57.40 $\pm$0.08} & \textbf{50.69 $\pm$0.45} & \textbf{90.54 $\pm$0.03} & \textbf{47.02 $\pm$0.87} \\
            \bottomrule
        \end{tabular}
    }
    \label{tab:comparison_with_sota}
\end{table*}

\subsection{Experimental Settings}
\label{sec:imple_detail}

\paragraph{Dataset}
We evaluate our proposed method on the MovieNet-SSeg dataset~\citep{movienet} that is a sub-dataset of MovieNet, containing 1,100 movies with 1.6M shots.
Note that only 318 out of 1,100 movies have scene boundary annotations, which are divided into 190, 64, and 64 movies for training, validation, and test split, respectively.
Following \cite{shotcol}, we use the entire 1,100 movies with no ground truth labels for the pre-training and fine-tune the model on the training split. The performance is measured on the test split.

\paragraph{Metric}
Following \cite{movienet}, we compare algorithms using Average Precision (AP) and mIoU that measures the averaged intersection over union (IoU) between predicted scene segments and their closest ground truth scene segments.
Also, we adopt F1 score and AUC-ROC as additional evaluation metrics.
Note that contrary to the previous works \citep{lgss,shotcol} that report recall, we use F1 score to consider for balanced comparison between precision and recall.
In addition, we report Meta-Sum metric inspired by the works \citep{uniter,value} for easy and straightforward comparison of algorithms.

\paragraph{Implementation details}
We employ ResNet-50~\citep{resnet} and Transformer~\citep{transformer} as the shot encoder and the contextual relation network, respectively.
For both pre-training and fine-tuning stages, we cross-validate the number of neighbor shots among $K = \{4, 8, 12, 16\}$ and $K=8$ is selected due to its good performance and computational efficiency.
In all experiments, given a pre-trained model, we fine-tune the model 5 times with different random seeds and report their average score and standard deviation.
More details are presented in supplementary material.

\subsection{Comparison with State-of-the-art Methods}
We compare our method, BaSSL, with 1) supervised methods including Siamese~\citep{siamese}, MS-LSTM~\citep{movienet} and LGSS~\citep{lgss}, 2) unsupervised methods including GraphCut~\citep{GraphCut}, SCSA~\citep{SCSA}, DP~\citep{DP}, StoryGraph~\citep{storygraph} and Grouping~\citep{grouping}, and 3) self-supervised methods including ShotCoL~\citep{shotcol}. Without fine-tuning on the downstream task, BaSSL can be seen as an unsupervised model in that it is trained to predict the pseudo-boundary by the PP task.
Table~\ref{tab:comparison_with_sota} summarizes comparison against competing methods.
BaSSL without fine-tuning shows competitive or outperforming performance based only on basic visual features compared to competing unsupervised methods;
note that the method, Grouping, leverages additional modalities (\eg, audio and transcripts).
Furthermore, fine-tuning BaSSL with ground-truth scene boundaries, AP is improved by 24.71\%p and BaSSL outperforms all other algorithms.
Finally, through longer pre-training (40 epochs), BaSSL surpasses the state-of-the-art method (\ie, ShotCoL) by a large margin (4.00\%p in AP).

\begin{table*}[!t]
    \caption{
        Average precision (AP) comparison with pre-training baselines.
        Note that SimCLR (NN) corresponds to our reproduced ShotCoL using SimCLR as the constrastive learning scheme. 
    }
    \vspace{0.1cm}
    \centering
    \scalebox{0.85}{
        \begin{tabular}{cl cc cc ccc}
            \toprule
            & \multirow{2}{*}{Method} & \multicolumn{2}{|c|}{Pre-training} & \multicolumn{2}{c|}{Transfer} & \multicolumn{3}{c}{Architecture of $f_{\text{CRN}}$ during fine-tuning} \\
            & & \multicolumn{1}{|c}{$f_{\text{ENC}}$} & \multicolumn{1}{c|}{$f_{\text{CRN}}$} & \multicolumn{1}{c}{$f_{\text{ENC}}$} & \multicolumn{1}{c|}{$f_{\text{CRN}}$} & \multicolumn{1}{c}{MLP} & MS-LSTM & Transformer \\
            \hline \hline
             
            \rowcolor[gray]{0.80}\multicolumn{9}{l}{\textit{\textbf{Supervised pre-training using image dataset}}} \\
            \hline
            M1 & \multicolumn{1}{l|}{ImageNet} & \checkmark & & \multicolumn{1}{|c}{\checkmark} & & \multicolumn{1}{|c}{43.12 $\pm$0.14} & 45.10 $\pm$0.55 & 47.13 $\pm$1.04 \\
            M2 & \multicolumn{1}{l|}{Places365} & \checkmark & & \multicolumn{1}{|c}{\checkmark} & & \multicolumn{1}{|c}{43.82 $\pm$0.10} & 45.87 $\pm$0.40 & 48.71 $\pm$0.50 \\
            
            \hline\hline    
            \rowcolor[gray]{0.80}\multicolumn{9}{l}{\textit{\textbf{Shot-level pre-training}}} \\
            \hline
            M3 & \multicolumn{1}{l|}{SimCLR (instance)} & \checkmark & & \multicolumn{1}{|c}{\checkmark} & & \multicolumn{1}{|c}{45.60 $\pm$0.07} & 49.09 $\pm$0.24 & 51.51 $\pm$0.31 \\
            M4 & \multicolumn{1}{l|}{SimCLR (temporal)} & \checkmark & & \multicolumn{1}{|c}{\checkmark} & & \multicolumn{1}{|c}{45.55 $\pm$0.11} & 49.24 $\pm$0.26 & 50.05 $\pm$0.78 \\
            M5 & \multicolumn{1}{l|}{SimCLR (NN)} & \checkmark & & \multicolumn{1}{|c}{\checkmark} & & \multicolumn{1}{|c}{45.99 $\pm$0.13} & 50.73 $\pm$0.19 & 51.17 $\pm$0.69 \\

            \hline\hline
            \rowcolor[gray]{0.80}\multicolumn{9}{l}{\textit{\textbf{Boundary-aware pre-training}}} \\
            \hline
            M6 &  \multicolumn{1}{l|}{BaSSL} & \checkmark & \checkmark & \multicolumn{1}{|c}{\checkmark} & & \multicolumn{1}{|c}{46.53 $\pm$0.11}  & 50.58 $\pm$0.14 & 50.82 $\pm$0.69 \\
            M7 & \multicolumn{1}{l|}{BaSSL} &\checkmark & \checkmark & \multicolumn{1}{|c}{\checkmark} & \checkmark & \multicolumn{1}{|c}{-} & - & 56.26 $\pm$0.04 \\
            \hline
            M8 & \multicolumn{1}{l|}{M5+M7} &\checkmark & \checkmark & \multicolumn{1}{|c}{\checkmark} & \checkmark & \multicolumn{1}{|c}{-} & - & 56.86 $\pm$0.01 \\ 
            \bottomrule 
        \end{tabular}
    }
    \vspace{-0.4cm}
    \label{tab:comp_pretraining}
\end{table*}

\subsection{Comparison with Pre-training Baselines}
We perform extensive experiments to compare BaSSL with the other pre-training baselines that learn shot-level representation by $f_{\text{ENC}}$.
In the experiments, we compare the following three types of pre-training approaches;
The first group (M1-2) trains $f_{\text{ENC}}$ using image-level supervision with object labels on ImageNet~\citep{imagenet} or place labels on Places365~\citep{places365}.
The second group (M3-5) trains $f_{\text{ENC}}$ through shot-level contrastive learning (\ie, SimCLR proposed by \cite{chen2020simple}) with different positive pair sampling strategies. Specifically, \emph{Instance} (M3) takes an instance of the center shot with different augmentation, \emph{Temporal} (M4) takes one randomly sampled neighbor shot as positive pair in local temporal window, and \emph{Nearest Neighbor (NN)} (M5) takes the most visually similar shot among the neighbor shots as positive pair, which is also known as ShotCoL~\citep{shotcol}.
The last group (M6-8) learns both $f_{\text{ENC}}$ and $f_{\text{CRN}}$ through boundary-aware pretext tasks proposed in this paper.
Given pre-trained representations of $f_{\text{ENC}}$, we train a video scene segmentation model with three different types of $f_{\text{CRN}}$ including MLP~\citep{shotcol}, MS-LSTM~\citep{movienet}\footnote{https://github.com/AnyiRao/SceneSeg/tree/master/lgss} and Transformer.
For fair comparison, all pre-training methods employ ResNet-50 as the shot encoder $f_{\text{ENC}}$ and we pre-train the models for 10 epochs.

In Table~\ref{tab:comp_pretraining}, we found the following observations.
First, when transferring pre-trained shot representation, employing MS-LSTM and Transformer as $f_{\text{CRN}}$ is more effective than using MLP, as they are favorably designed to capture contextual relation between shots (see M1-6).
Second, BaSSL (M7) outperforms all competing baselines (M1-5) through learning contextual representation during pre-training.
Also, it turns out that transferring the representation through $f_{\text{CRN}}$ is important for the boundary detection task where it leads to a performance gain of 5.44\%p in AP (see M6-7).
Finally, learning shot-level and contextual representations is complementary to each other;
that is, incorporating ShotCoL (M5) and our framework (M7) provides further improved performance (M8).

\subsection{Ablation Studies}

\begin{table*}[!t]
    \caption{
        Ablation study on varying combinations of pretext tasks for pre-training.
        The best scores are highlighted in bold.
    }
    \vspace{0.1cm}
    \centering
    \begin{adjustbox}{width=1.0\textwidth}
        \begin{tabular}{c cccc c ccccc}
            \toprule
            \textit{} & \multicolumn{4}{c}{Pretext Tasks} & \multicolumn{1}{c}{} & \multicolumn{5}{c}{Evaluation Metric} \\
            \cline{1-5} \cline{7-11}
             & \begin{tabular}[c]{@{}c@{}}SSM\end{tabular} & \begin{tabular}[c]{@{}c@{}}CGM\end{tabular} & \begin{tabular}[c]{@{}c@{}} PP \end{tabular} & \begin{tabular}[c]{@{}c@{}}MSM\end{tabular} & & AP & mIoU & AUC-ROC & F1 & Sum \\
            
            \hline \hline
            
            P1 & \checkmark &  &  &  &  & 42.57 $\pm$0.29 & 40.12 $\pm$0.50 & 84.11 $\pm$0.15 & 30.83 $\pm$0.79 & 197.63 \\
            P2 &  & \checkmark &  &  &  & 36.76 $\pm$0.02 & 40.59 $\pm$0.18 & 82.06 $\pm$0.04 & 30.94 $\pm$0.32 & 190.35 \\
            P3 &  &  & \checkmark &  & & 36.55 $\pm$0.04 & 39.58 $\pm$0.05 & 81.36 $\pm$0.03 & 29.96 $\pm$0.04 & 187.45 \\
            P4 &  &  &  & \checkmark &  & 13.33 $\pm$0.23 & 29.80 $\pm$0.39 & 64.65 $\pm$0.98 & 18.68 $\pm$0.39 & 126.45 \\

            \hline
            P5 & \checkmark & \checkmark &  & &  & 55.77 $\pm$0.05 & 48.19 $\pm$0.21 & 90.19 $\pm$0.03 & 43.17 $\pm$0.39 & 237.32 \\
            P6 & \checkmark &  & \checkmark &  &  & 56.04 $\pm$0.08 & 49.00 $\pm$0.16 & 90.13 $\pm$0.02 & 44.74 $\pm$0.29 & 239.91 \\
            P7 &  & \checkmark & \checkmark &  &  & 38.09 $\pm$0.03 & 41.25 $\pm$0.10 & 82.85 $\pm$0.01 & 32.24 $\pm$0.24 & 195.43 \\
            P8 & \checkmark &  &  & \checkmark &  & 54.39 $\pm$0.07 & 47.54 $\pm$0.18 & 89.72 $\pm$0.03 & 42.48 $\pm$0.22 & 234.13 \\
            P9 &  & \checkmark &  & \checkmark &  & 39.49 $\pm$0.04 & 41.71 $\pm$0.12 & 83.27 $\pm$0.02 & 32.85 $\pm$0.20 & 197.32 \\
            P10 &  &  & \checkmark & \checkmark &  & 38.53 $\pm$0.07 & 40.85 $\pm$0.15  &  82.78 $\pm$0.04  & 31.47 $\pm$0.16 & 193.63 \\

            \hline
            P11 &  & \checkmark & \checkmark & \checkmark &  & 41.02 $\pm$0.07 & 40.89 $\pm$0.10  & 83.79 $\pm$0.02 & 31.53 $\pm$0.18 & 197.23 \\ 
            P12 & \checkmark &  & \checkmark & \checkmark &  & 56.10 $\pm$0.08 & 49.10 $\pm$0.17 & 90.09 $\pm$0.03 & 45.42 $\pm$0.30 & 240.71 \\
            P13 & \checkmark & \checkmark &  & \checkmark & & 56.20 $\pm$0.06 & 48.00 $\pm$0.17 & 90.13 $\pm$0.01 & 43.24 $\pm$0.27 & 237.57 \\
            P14 & \checkmark & \checkmark & \checkmark &  &  & \textbf{56.26 $\pm$0.02} & 48.42 $\pm$0.33 & 90.25 $\pm$0.01 & 43.98 $\pm$0.58 & 238.91 \\

            \hline
            P15 & \checkmark & \checkmark & \checkmark & \checkmark &  & \textbf{56.26 $\pm$0.04} & \textbf{49.50 $\pm$0.11} & \textbf{90.27 $\pm$0.02} &\textbf{45.70 $\pm$0.24} & \textbf{241.73} \\
            \bottomrule
        \end{tabular}
    \end{adjustbox}
    \vspace{-0.2cm}
    \label{tab:ablation_pretrain_task}
\end{table*}

\begin{table*}[!t]
	\centering
	\caption{
		Ablations to check the impact of pseudo-boundary discovery strategies, the number of neighboring shots ($K$) and longer pre-training.
		The best scores are in bold.
	}
	\vspace{-0.05cm}
	\begin{minipage}[t]{0.37\linewidth}
		\centering
		\subfloat[Performance comparison depending on the pseudo-boundary discovery methods. \label{tab:ablation:pseudo_bd}]{
			\scalebox{0.9}{
			\begin{tabu}{l|c}
				\toprule
				Pseudo-boundary & AP \\
				\hline \hline
				Random      & 46.64 $\pm$0.37 \\
				Fixed       & 49.53 $\pm$0.32 \\
				Synthesized & 54.61 $\pm$0.03 \\
				\textbf{DTW (ours)} & \textbf{56.26 $\pm$0.04} \\
				\tabucline[1pt]{-}
			\end{tabu}
			}
		}
	\end{minipage}
	\begin{minipage}[t]{0.33\linewidth}
		\centering
		\subfloat[Performance comparison when varying the number of neighbor shots. \label{tab:ablation:num_neighbor}]{
			\scalebox{0.9}{
			\begin{tabu}{c|c}
				\toprule
			    \# Neighbors & AP \\
				\hline \hline
				4  & 55.98 $\pm$0.10 \\
				8  & 56.26 $\pm$0.04 \\
				\textbf{12} & \textbf{56.29 $\pm$0.03} \\
				16 & 55.31 $\pm$0.04 \\
				\tabucline[1pt]{-}
			\end{tabu}
			}
		}
	\end{minipage}
	\begin{minipage}[t]{0.28\linewidth}
		\centering		
		\subfloat[Performance comparison with respect to the number of pre-training epochs. \label{tab:ablation:epoch}] {
		\centering
		\scalebox{0.9}{
		\begin{tabu}{c|c}
			\toprule
			Epochs & AP \\
			\hline \hline
			10 & 56.26 $\pm$0.04 \\
			20 & 56.74 $\pm$0.04 \\
			30 & 56.74 $\pm$0.07 \\
			\textbf{40} & \textbf{57.40 $\pm$0.08} \\
			50 & 57.15 $\pm$0.08 \\
			\tabucline[1pt]{-}
		\end{tabu}
		}
	}
	\end{minipage}
	\vspace{-0.2cm}
	\label{tab:additional_ablations}
\end{table*}


\paragraph{Impact of individual pretext tasks}
We investigate the contribution of individual pretext tasks.
In this experiment, we train models by varying the combinations of the pretext tasks.
From Table~\ref{tab:ablation_pretrain_task}, we can obtain following two observations.
First, when training a model with a single pretext task (P1-4), the MSM task leads to the worst performance compared to the others.
This indicates that boundary-aware pretext tasks (\ie, SSM, CGM and PP) to learn contextual relation between shots is indeed important for video scene segmentation.
Second, the more pretext tasks we include during pre-training, the better the performance is, and the best performance is obtained when using all tasks (P15). 
This means all tasks are complementary to each other, contributing to performance gain.

\paragraph{Psuedo-boundary discovery method}
To check the effectiveness of DTW-based pseudo-boundary discovery, we train three models with different pseudo-boundary decision strategies---1) \emph{Random} defining one randomly sampled shot in the input sequence as a pseudo-boundary, 2) \emph{Fixed} always taking the center shot as a pseudo-boundary, and 3) \emph{Synthesized}, inspired by \cite{bsp}, synthesizing the input sequence by concatenating two sub-sequences sampled from different movies and using the last shot of the first sub-sequence as a pseudo-boundary. 
Table~\ref{tab:additional_ablations}(a) summarizes the results.
\emph{Random} and \emph{Fixed} pseudo-boundaries hinder the learning and degenerate the boundary detection performance.
It is notable that BaSSL with \emph{Synthesized} pseudo-boundaries also outperforms the pre-training baselines in Table~\ref{tab:comp_pretraining}, which shows the effectiveness of our framework and importance of pre-training contextualized representation.
Finally, adopting DTW to find pseudo-boundaries achieves the best performance.

\paragraph{Hyperparameters}
We analyze the impact of two key hyperparameters: 1) the number of neighbor shots $K$ and 2) pre-training epochs.
Table~\ref{tab:additional_ablations}(b) shows that we achieve higher performance with more neighbor shots, saturating around $K = 12$.
Table~\ref{tab:additional_ablations}(c) shows the impact of longer pre-training.
We find that performance increases until certain numbers (\ie, 40 epochs) and decrease afterward.
We conjecture that this is partly due to overfitting to noise from incorrect pseudo-boundaries.

\section{Conclusion}
We present BaSSL, a novel self-supervised framework for video scene segmentation, especially designed to learn contextual relationship between shots. 
Through the pseudo-boundary discovery, we can define and conduct boundary-aware pretext tasks that encourage the model to learn the contextual relational representation and a capability of capturing transitional moments.
Comprehensive experiments demonstrate the effectiveness of our framework and we achieve outstanding performance in the MovieNet-SSeg dataset.

\subsubsection*{Acknowledgments}
We would like to acknowledge and thank Brain Cloud Team at Kakao Brain for their support.



\bibliography{paper}
\bibliographystyle{iclr2022_conference}

\newpage
\appendix

\section{Additional Implementation Details}
\label{sec:add_detail}
Additional details of the shot encoder (\ie, ResNet-50) and the contextual relation network (\ie, Transformer) are as follows.
For the shot encoder, each shot is given by three key-frames (\ie, $N_k=3$) and a shot encoding $\mathbf{e}$ is given by the averaged feature after inferring individual three key-frames using ResNet-50;
note that, to speed up the training, we use randomly sample one key-frame out of three during the pre-training.
For Transformer, the hyperparameters are set to ($L=2, H=768, A=8$) where $L$, $H$ and $A$ mean the number of stacked transformer blocks, the dimension of hidden activation and the number of attention heads, respectively.
We apply the Dropout technique~\citep{dropout} on hidden states and attention weights with a probability of 10\% and use GELU~\citep{gelu} as an activation function.

For data augmentation of key-frames in a shot, we adopt PyTorch's torchvision package.
Given a sequence of shots, we apply random crop (with resize), random flip, random color jitter and random Gaussian blur.
In detail, firstly, the cropping is performed with a random size (\ie, scales between [0.14, 1.0] of the original size) and a random aspect ratio (between 3/4 to 4/3), and then the cropped one is resized to (224,224).
Secondly, we apply a random horizontal flip with a probability of 50\%.
Thirdly, as a color augmentation, we perform a random color jitter (with a probability of 80\%) and a random color dropping to gray scale (with a probability of 20\%).
The color strength parameters for jittering are set to $\{\text{brightness}, \text{contrast}, \text{saturation}, \text{hue}\}=(0.2, 0.2, 0.2, 0.05)$.
Finally, Gaussian blur is applied with a probability of 50\% where a standard-deviation of spatial kernel is set to [0.1, 2.0].
Note that the same augmentations are applied to all key-frames in the input sequence $\ma{S}_n$ of shots while different color jittering is applied on individual shots.
Also, for $\ma{S}_n^{\text{slow}}$, we perform a different augmentation compared to that applied on $\ma{S}_n$.

During the pre-training stage, the model parameters are randomly initialized and then trained using the proposed pretext tasks.
We use LARS~\citep{LARS} to learn the model (except for parameters of bias and Batch-Normalization) with a mini-batch of 256 shot sequences, a base learning rate of 0.3, momentum of 0.9, weight decay of $10^{-6}$ and trust coefficient of 0.001.
We pre-train the model for 10 epochs with a linear warm-up strategy for 1 epoch (\ie, 10\% of whole training epochs) followed by learning rate decaying with a cosine schedule.
The temperature $\tau$ in Eq. (\ref{eq:nce_loss}) is set to 0.1.
Using 16 V100 GPUs with mixed precision training, it takes less than 2 days for pre-training.

In the fine-tuning stage, we initialize the parameters of the shot encoder and the contextual relation network by that of the pre-trained ones.
However, we freeze the parameters of the shot encoder following \cite{shotcol}.
We fine-tune the contextual relation network and the scene boundary detection head for 20 epochs using Adam~\citep{Adam} with a learning rate of $10^{-5}$ and a mini-batch of 1024 training examples.
The learning rate is decayed with a cosine schedule without a warm-up stage.

\section{Comparison with Shot-level Self-supervised Learning}
\begin{figure}[!t]
	\centering
	\scalebox{0.9}{
		\includegraphics[width=\linewidth]{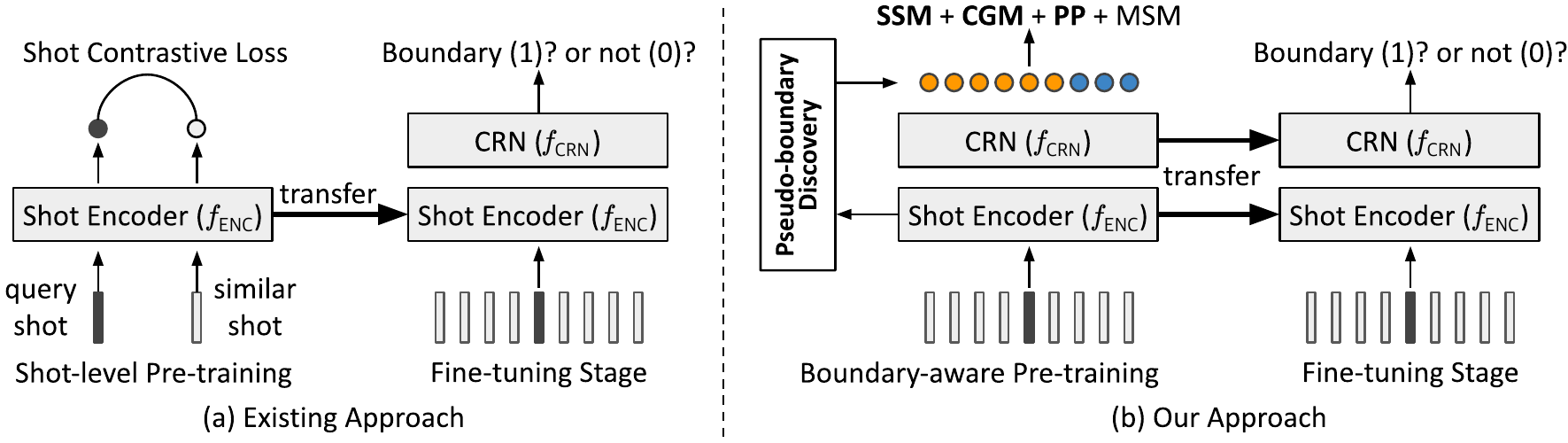}
	}
	\vspace{-0.2cm}
	\caption{
		Comparison between existing approaches and ours for video scene segmentation.
		The existing approach focuses only on learning shot-level representation given by shot encoder ($f_{\text{ENC}}$).
		In contrast, our boundary-aware pre-training method focuses on learning contextual representation by taking neighbor shots into account.
		Thus, our method can learn both the shot encoder ($f_{\text{ENC}}$) and the contextual relation network (CRN; $f_{\text{CRN}}$) and transfer their parameters during the fine-tuning stage.
	}
	\label{fig:algo_comp}
\end{figure}

As mentioned in the main paper, our approach is distinguishable from the shot-level pre-training approach~\citep{shotcol} in that the objectives used in our approach (BaSSL) is to learn contextual representations by taking neighbor shots into account. Figure~\ref{fig:algo_comp} provides a clear summary of comparison between shot-level pre-training and our boundary-aware pre-training, BaSSL.
Firstly, shot-level pre-training takes a pair of two shots as an input while BaSSL takes a sequence of shots.
Secondly, shot-level pre-training aims to train shot encoder ($f_{\text{ENC}}$) only, while BaSSL trains both the shot encoder and the contextual relation network ($f_{\text{ENC}}$ and $f_{\text{CRN}}$). In contrast to the shot-level pre-training that requires to train $f_{\text{CRN}}$ from scratch during the fine-tuning stage, BaSSL benefits from weight transfer by pre-training the parameters of $f_{\text{CRN}}$ with large-scale in-domain data in advance. Note that the results (M6-7) in Table~\ref{tab:comp_pretraining} show that the weight transfer of $f_{\text{CRN}}$ is important to improve the video scene segmentation performance.
Finally, the contrastive learning objective in shot-level pre-training drives the representations of two shots (query and positive) to be close to each other, whereas Shot-Scene Matching objective in our approach performs the same task but with a shot (query) and its associated scene (positive; a sequence of shots).
The Table~\ref{tab:checklist} summarizes the aforementioned comparisons.

\begin{table*}[!t]
    \caption{
        Comparison between the shot-level pre-training and the proposed pre-training approaches.
    }
    \vspace{0.1cm}
    \centering
    \scalebox{0.90}{
        \begin{tabular}{lcc}
        \toprule
        Check List & Shot-level Pre-training & Boundary-aware Pre-training \\
        \hline \hline
        Network architecture & $f_{\text{ENC}}$ & $f_{\text{ENC}}$ + $f_{\text{CRN}}$ \\
        Training input & a pair of shots (\#shots: 2) & a sequence of shots (\#shots: 2K+1) \\
        Weights transferable for $f_{\text{ENC}}$? & yes & yes \\
        Weights transferable for $f_{\text{CRN}}$? & no & yes \\
        Positive pair in contrastive learning & shot-shot & shot-scene \\
        
        \bottomrule
        \end{tabular}
    }
    \label{tab:checklist}
\end{table*}

\begin{table*}[!t]
    \caption{
        Comparison of existing video scene segmentation datasets. Note that we brought the table from ~\citep{lgss} with an update on the MovieNet-SSeg dataset.
    }
    \vspace{0.1cm}
    \centering
    \scalebox{0.85}{
        \begin{tabular}{lccccc}
        \toprule
        Dataset  & \#Video & \#Scene & \#Shot & Time (h) & Source   \\
        \hline \hline
        BBC~\citep{siamese}   & 11      & 670     & 4.9K  & 9      & Documentary  \\
        OVSD~\citep{OVSD}     & 21      & 300     & 10K & 10     & MiniFilm  \\
        \hline
        MovieNet-SSeg~\citep{movienet} & 318 & 42K & 500K & - & Movies   \\
        MovieNet~\citep{movienet} & 1,100 & - & 1.6M & - & Movies   \\
        \bottomrule
        \end{tabular}
    }
    \label{tab:dataset_statistics}
\end{table*}

\begin{table}[!t]
    \caption{
        Comparison between our method and shot-level pre-training baselines on BBC and OVSD datasets. The numbers mean AP and the best model is highlighted in bold.
    }
    \vspace{0.1cm}
    \centering
    \begin{adjustbox}{width=0.8\textwidth}
        \begin{tabular}{l cccc}
        \toprule
         Model & SimCLR (instance) & SimCLR (temporal) & SimCLR (NN) & BaSSL  \\
         \hline \hline
        BBC  & 32.34 & 34.18 & 32.92 & \textbf{39.98} \\
        OVSD & 25.45 & 24.92 & 25.02 & \textbf{28.68} \\
        \bottomrule
        \end{tabular}
    \end{adjustbox}
    \vspace{-0.2cm}
    \label{tab:bbc_ovsd_result}
\end{table}

\section{Results on additional datasets}
Table~\ref{tab:dataset_statistics} shows the data statistics of different video scene segmentation datasets. We found the limited number of datasets that provide the scene boundary annotations and, as far as we know, the MovieNet-SSeg~\citep{movienet} is the largest-scale video scene segmentation dataset.
We further compare BaSSL with shot-level pre-training baselines on two additional datasets---BBC~\citep{siamese} and OVSD~\citep{OVSD}. 
Note that the training and test splits are not available and the dataset size is extremely limited (11 and 21 videos in BBC and OVSD, respectively); in addition, 2 out of 21 videos in OVSD is not available.
Thus, we infer predictions using models trained on MovieNet-SSeg without fine-tuning on BBC and OVSD. The results are summarized in Table~\ref{tab:bbc_ovsd_result}. 
The result shows the superiority of our method compared to shot-level pre-training baselines.

\section{Algorithm for Pseudo-boundary Discovery}
In this section, we describe the details of pseudo-boundary discovery method applying DTW on $\mathbf{S}_n$ and $\mathbf{S}_n^{\text{slow}}$.
In practice, $\mathbf{S}_n$ is given as a mini-batch resulting in a tensor with a shape of ($B$, $S$, $N_k$, $C$, $H$, $W$) where individuals mean the batch size, the number of shots in $\mathbf{S}_n$ (\ie, $2K+1$), the number of key-frames in a shot, channels, frame height and frame width, respectively.
Then, we obtain $\mathbf{S}_n^{\text{slow}} \in \mathbb{R}^{B\times 2\times N_k\times C\times H\times W}$ that is composed of the first and last shots in $\mathbf{S}_n$.
We apply two different augmentation functions into key-frames in $\mathbf{S}_n$ and $\mathbf{S}_n^{\text{slow}}$, respectively.
Next, we compute encoded representation of shots from $\mathbf{S}_n$ and $\mathbf{S}_n^{\text{slow}}$ using $f_{\text{ENC}}$.
Note that during the pre-training stage, we randomly sample one key-frame among $N_k$ candidates in a shot and then reshape the input tensor as ($B$*$S$, $C$, $H$, $W$) or ($B$*2, $C$, $H$, $W$) to be forwarded by the shot encoder $f_{\text{ENC}}$;
thus the tensor shape of the encoded shot representation is given by ($B$, $S$, $D_e$) or ($B$, 2, $D_e$) after apply reshaping, where $D_e$ means the dimension of encoded feature.
Finally, given two sequences of encoded representation for $\mathbf{S}_n$ and $\mathbf{S}_n^{\text{slow}}$, DTW provides two sub-sequences $\ma{S}^\text{left}_n$ and $\ma{S}^\text{right}_n$ and a pseudo boundary shot $\mathbf{s}_{n+b^{*}}$.
The algorithm~\ref{algo:dtw} illustrates the details.
In addition, to demonstrate the simplicity of the alignment computation using DTW, we include the PyTorch code in Listing~\ref{lst:dtw_code}. The implementation of DTW can be done in 5 lines of python code using \textit{tslearn} package.

\definecolor{light-gray}{gray}{0.5}  
\begin{algorithm}[!t]
	\caption{DTW-based pseudo-boundary discovery} 
	\begin{algorithmic}[1]
	    \State \textbf{Input}: Shot encoder $f_{\text{enc}}$, contextual relation network $f_{\text{CRN}}$, and an input shot sequence $\mathbf{S}_n=\{\mathbf{s}_{n-K}, ..., \mathbf{s}_n, ..., \mathbf{s}_{n+K}\}$ centered at $n^{\text{th}}$ shot $\mathbf{s}_n$ with neighbor size $K$, two image augmentation functions $\lambda_{\text{aug}}^{1}, \lambda_{\text{aug}}^{2}$.
	    \State $(\ma{E}_n, \ma{E}^\text{slow}_{n}) \leftarrow$ ([], [])
	    \For {$i$ = $n-K$ \textbf{to} $n+K$ }
	        \State $\mathbf{e}_i \leftarrow f_{\text{ENC}}(\lambda_{\text{aug}}^{1}(\mathbf{s}_i))$ \textcolor{light-gray}{~~~// extract shot-level representations for all shots}
	        \State $\mathbf{E}_{n} \leftarrow \{ \mathbf{E}_{n}; \mathbf{e}_i \}$
	        \textcolor{light-gray}{~~~// append}
	    \EndFor
	    \For {$i$ \textbf{in} \{$n-K$, $n+K$\} }
	        \State $\mathbf{e}_i \leftarrow f_{\text{ENC}}(\lambda_{\text{aug}}^{2}(\mathbf{s}_i))$ \textcolor{light-gray}{~~~// extract shot-level representations for slow sequence}
	        \State $\ma{E}^\text{slow}_{n} \leftarrow \{ \ma{E}^\text{slow}_{n}; \mathbf{e}_i \}$
	        \textcolor{light-gray}{~~~// append}
	    \EndFor
	    \State $\mathbf{S}_n^{\text{left}}$, $\mathbf{S}_n^{\text{right}}$, $b^{*}$ $\leftarrow$ \text{DTW}($\ma{E}_n, \ma{E}^\text{slow}_{n}$) \textcolor{light-gray}{~~~// apply dynamic time warping}
	    
	    \State \textbf{Output}: Two continuous non-overlapping sub-sequences $\mathbf{S}^\text{left}_n$ and $\mathbf{S}^\text{right}_n$ and a pseudo boundary shot $\mathbf{s}_{n+b^{*}}$.
	\end{algorithmic}
	\label{algo:dtw}
\end{algorithm}

\begin{lstlisting}[language=Python,basicstyle=\footnotesize\ttfamily,caption={PyTorch code for alignment computation using DTW given two sequences. The \textit{tslearn} package is used for DTW path calculation.},label={lst:dtw_code},captionpos=b,float=tp,floatplacement=tbp]
from tslearn import metrics
import numpy as np
def compute_dtw_path(self, seq_1, seq_2):
    """
    Input:
        seq_1: sparse shots embedding, shape = torch.Size([2, dim])
        seq_2: dense shots embedding, shape = torch.Size([N, dim]), N > 2
    Output:
        dtw_path: output of DTW algorithm, shape = torch.Size([N, dim])
    """
    cost = (1-torch.bmm(seq_1, seq_2.transpose(1, 2))).numpy()
    dtw_path = []
    for bsz in range(cost.shape[0]):
        _path, _ = metrics.dtw_path_from_metric(cost[bsz], metric="precomputed")
        dtw_path.append(np.asarray(_path)) # torch.Size([N, dim])
    return dtw_path
\end{lstlisting}

\begin{table}[!t]
    \caption{
        Scene clustering quality measured by normalized mutual information (NMI) metric.
    }
    \vspace{0.1cm}
    \centering
    \begin{adjustbox}{width=0.95\textwidth}
        \begin{tabular}{l cccc}
        \toprule
        \multirow{2}{*}{Model} & \multicolumn{3}{c}{Scene Length} & \multirow{2}{*}{$\Delta$ $\downarrow$ (Short $\rightarrow$ Long)} \\
        \cline{2-4}
         & Short ($N_c$=8) & Medium ($N_c$=16) & Long ($N_c$=32) &  \\
         \hline \hline
        ImageNet & 67.50 & 61.60 & 56.25 & -16.67\% \\
        SimCLR (temporal) & 82.40 & 81.65 & 78.99 & -4.14\% \\
        SimCLR (NN) & 83.54 & 83.17 & 81.25 & -2.75\% \\
        \hline
        BaSSL (ours) & \textbf{86.22} & \textbf{86.72} & \textbf{85.63} & \textbf{-0.68\%} \\
        \bottomrule
        \end{tabular}
    \end{adjustbox}
    \vspace{-0.2cm}
    \label{tab:nmi}
\end{table}

\begin{table}[!t]
    \caption{
        Ablation study on the combination of boundary-aware pretext tasks measured by NMI.
    }
    \vspace{0.1cm}
    \centering
    \begin{adjustbox}{width=0.45\textwidth}
        \begin{tabular}{l cc}
        \toprule
        Pretext Tasks & NMI & Gain ($\Delta$\%) \\
        \hline \hline
        SSM & 85.48 & 0.00\% \\
        SSM+MSM & 85.64 & +0.19\% \\
        SSM+MSM+CGM & 85.93 & +0.33\% \\
        SSM+MSM+CGM+PP & 86.71 & +0.91\% \\
        \bottomrule
        \end{tabular}
    \end{adjustbox}
    \vspace{-0.2cm}
    \label{tab:nmi_pretext}
\end{table}

%
\section{Measuring Representation Quality at Pre-training Stage}

The normalized mutual information (NMI) is a metric for clustering algorithms (\eg, K-Means), which measures the clustering quality. 
Since clustering with good representations forms clear boundaries between different classes, NMI can be considered as a proxy to measure the quality of our pre-trained models.
Specifically, we randomly sample 100 scenes from the test split of MovieNet-SSeg while we vary the length of scenes $N_c \in \{8, 16, 32\}$.
Then, we perform K-Means clustering on $N_c$ $\times$ 100 shot representations extracted by the pre-trained model with the number of classes $K$=100. 
This intends to form a single cluster for each scene, assuming that high-quality representation for movie scene segmentation would locate the shot embeddings within the same scene close to each other.
Considering the randomness in the K-Means clustering and scene sampling, we report the averaged score from five trials.

In Table~\ref{tab:nmi}, we compare the NMI score between different pre-trained models;
SimCLR (NN) is our SimCLR version implementation of ShotCoL. 
The result shows that BaSSL outperforms the shot-level pre-training baselines and the model pre-trained using ImageNet dataset.
With respect to different scene lengths ($N_c$; the number of shots included in a single scene), we found our BaSSL is more robust than the other baselines.
Since the visual diversity across the shots increases as the scenes become longer ($N_c$=8 $\rightarrow$ 32), it is natural that the NMI score for each baseline is degraded.
However, it is remarkable that, by increasing the number of shots from 8 to 32, the performance of BaSSL drops only -0.68\% while the other baselines suffer from severe degradation.
This demonstrates the effectiveness of BaSSL in maximizing intra-scene similarity.

In addition, we perform ablation study of our algorithm by adding pretext tasks one by one, and measure the corresponding NMI scores.
The result in Table~\ref{tab:nmi_pretext} shows that better NMI score is achieved as more pretext tasks are combined together. This tendency is also observed in our ablation in Table~\ref{tab:ablation_pretrain_task}, which indicates the NMI score of pre-trained models is highly correlated with the final performance after the fine-tuning.

\section{Qualitative Analysis}

\begin{figure}[!t]
	\centering
	\scalebox{1}{
		\includegraphics[width=\linewidth]{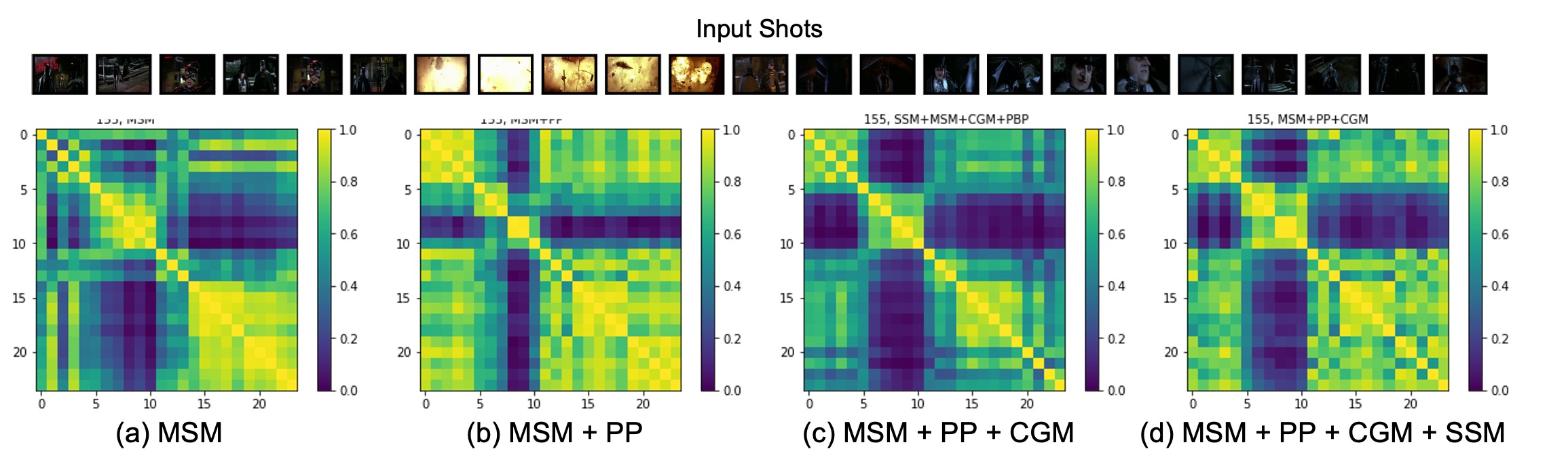}
	}
	\vspace{-0.7cm}
	\caption{
		Visualization of similarity (below) between shot representations in randomly sampled consecutive shots (above). We observe that the shot representations are clearly clustered  as adding pretext tasks one by one.
	}
	\label{fig:shot_sim}
\end{figure}

\paragraph{Visualization of similarities between consecutive shots}
To qualitatively check the effect of individual pretext tasks, we visualize the matrix of cosine similarity between shot representations from the randomly sampled 16 consecutive shots in Figure~\ref{fig:shot_sim}.
The shot representations are computed by models without the fine-tuning in order to solely focus on the behavior of each objective at the pretraining stage. 
When the MSM is used only, approximately three clusterings are shown, but similarity around boundaries is smoothed. 
Next, when we add PP, dissimilarities around the boundaries are to be sharpened. Then, with additional CGM, the clusters are more clearly obtained. 
Finally, adding SSM makes the similarity of shots within the same cluster higher (i.e., more yellow ones).

\begin{figure}[!t]
	\centering
	\scalebox{1.0}{
		\includegraphics[width=\linewidth]{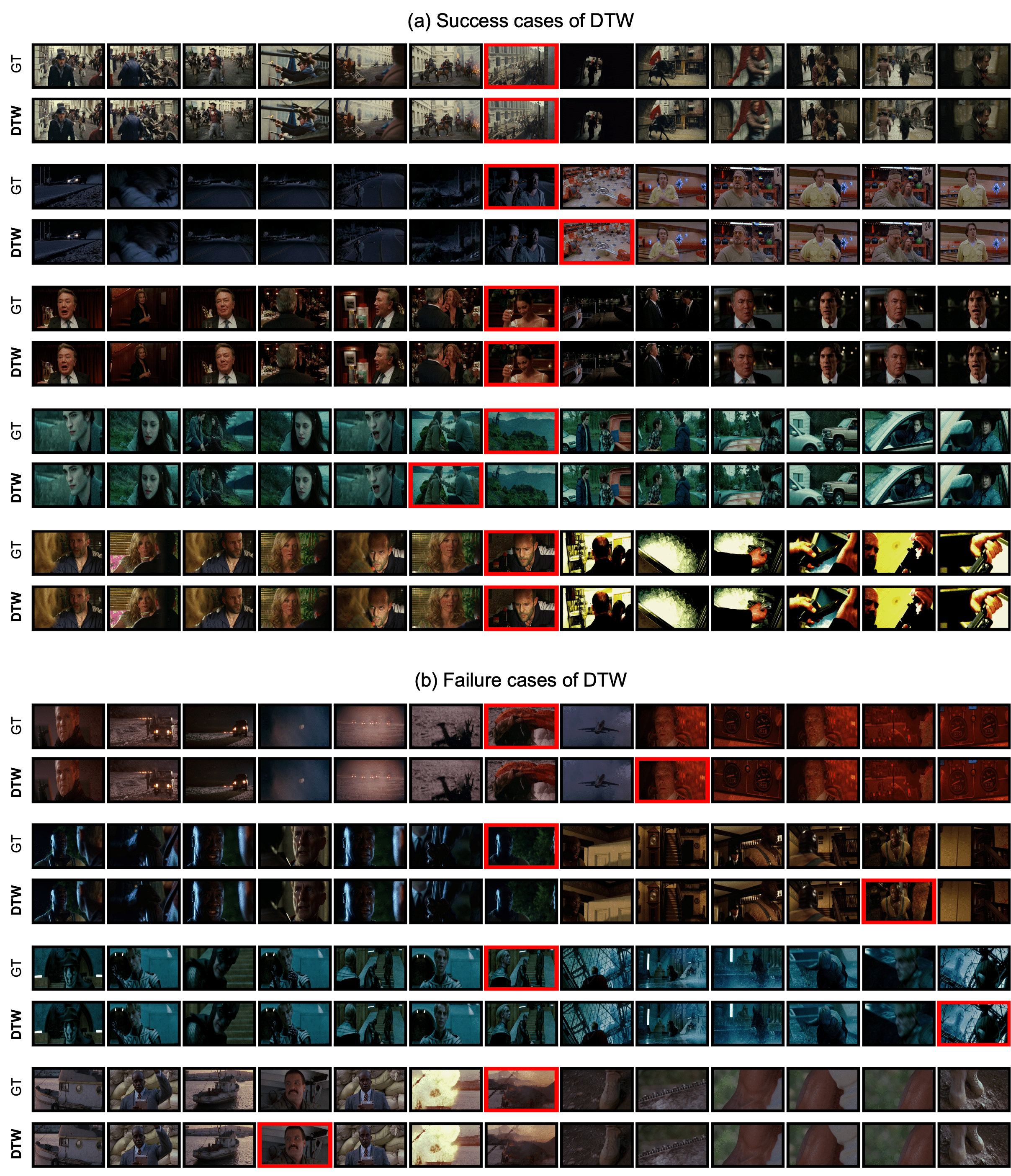}
	}
	\vspace{-0.5cm}
	\caption{
	    Comparison between the ground truth scene boundaries and the discovered pseudo-boundaries based on the DTW algorithm.
	    The examples are sampled from the MovieNet-SSeg dataset.
		All boundary shots are highlighted in red. 
	}
	\label{fig:example_gt}
\end{figure}

\paragraph{Pseudo-boundaries}
We compare the quality of discovered pseudo-boundaries with the ground truth scene boundaries in Figure~\ref{fig:example_gt}. In most cases, we observe the pseudo-boundaries identified by the DTW algorithm are successfully located in close distance with the ground truth ones. This result validates our idea considering the problem of discovering pseudo-boundary as a temporal alignment problem between two sequences with different frequencies ($\ma{S}_n$ and $\ma{S}_n^\text{slow}$). At the same time, we illustrate the failure cases. Although discovered pseudo-boundary does not match the ground truth in this case, we figure the determined boundary is not always arbitrary. For example, the mismatch is often caused by the noise existing in the ground truth (see the first row in the failure cases). On the other hand, in case all shots are visually similar (see the third row in the failure cases), the DTW solely relying on the visual modality fails to find the correct boundary.

\begin{figure}[!t]
	\centering
	\scalebox{1.0}{
		\includegraphics[width=\linewidth]{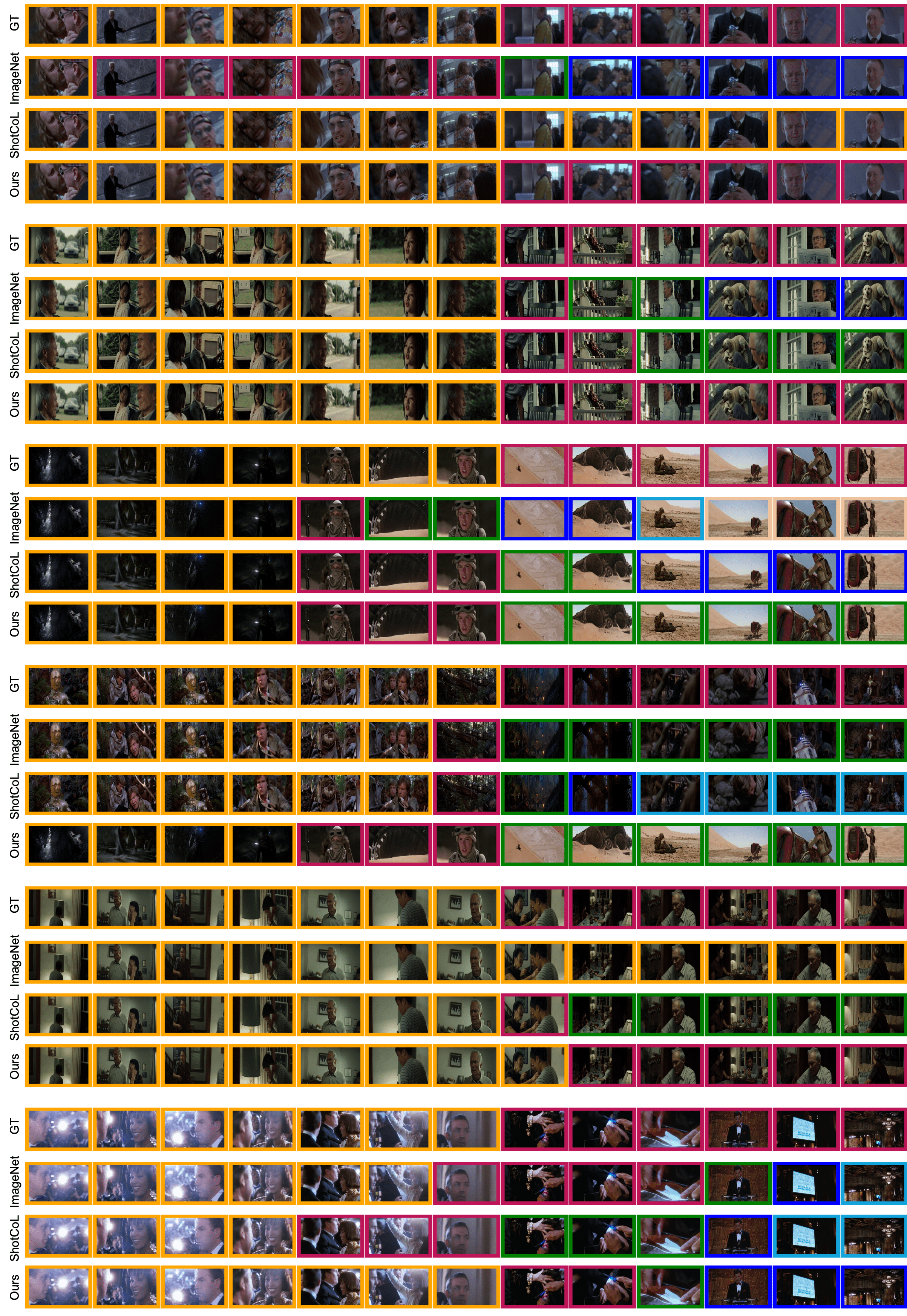}
	}
	\vspace{-0.5cm}
	\caption{
		Comparison of boundary detection results from three pre-training approaches: ImageNet pre-trained ResNet, ShotCoL, and BaSSL. 
		The first row shows the reference that is composed of two adjacent scenes divided by the ground truth boundary.
		We visualize the shots that are assigned to the same scene segments with the same colored border.
	}
	\label{fig:example_prediction_resized}
\end{figure}

\paragraph{Predicted scene boundaries}
The figure~\ref{fig:example_prediction_resized} illustrates the scene boundary predictions of different models. Comparing with the baselines, we observe that our approach, BaSSL, shows qualitatively better result for video scene segmentation. On the other hand, we observe the over-segmentation issue in many cases using any competing methods (including ours). Our finding implies that achieving the highest recall only does not guarantee the highest performance in practice. We reckon that further studies on this over-segmentation problem would be a highly important topic when it comes to real-world application.

\end{document}

%% file: math_commands.tex
\usepackage{amsmath,amsfonts,bm}

\def\eqref#1{equation~\ref{#1}}

\def\1{\bm{1}}

\DeclareMathAlphabet{\mathsfit}{\encodingdefault}{\sfdefault}{m}{sl}
\SetMathAlphabet{\mathsfit}{bold}{\encodingdefault}{\sfdefault}{bx}{n}

\DeclareMathOperator*{\argmax}{arg\,max}